\documentclass[letterpaper, 10 pt, conference]{ieeeconf}  

\IEEEoverridecommandlockouts                              

\usepackage{graphicx} 
\usepackage{amsmath} 
\usepackage{amssymb}  
\usepackage{multirow}
\usepackage{caption}
\usepackage{subcaption}
\usepackage{booktabs}
\let\labelindent\relax
\usepackage[inline]{enumitem}
\usepackage{hyperref}

\title{\LARGE \bf
Evaluating Vision Transformer Methods \\for Deep Reinforcement Learning from Pixels
}

\author{Tianxin Tao$^{*}$ \hspace{18pt} Daniele Reda$^{*}$ \hspace{18pt} Michiel van de Panne
\thanks{*The authors contribute equally}
\thanks{Tianxin Tao, Daniele Reda and Michiel van de Panne are with Department of Computer Science, University of British Columbia, Vancouver, BC, Canada. Email: \texttt{\{taotianx, dreda, van\}@cs.ubc.ca.}}%
}

\begin{document}

\maketitle
\thispagestyle{empty}
\pagestyle{empty}

\begin{abstract}

Vision Transformers (ViT) have recently demonstrated the significant potential of transformer architectures 
for computer vision. To what extent can image-based deep reinforcement learning also benefit from ViT architectures,
as compared to standard convolutional neural network (CNN) architectures?
To answer this question, we evaluate ViT training methods for image-based reinforcement learning~(RL) control tasks
and compare these results to a leading convolutional-network architecture method, RAD.
For training the ViT encoder, we consider several recently-proposed self-supervised losses that are treated as auxiliary tasks, as well as a baseline with no additional loss terms.
We find that the CNN architectures trained using RAD still generally provide superior performance.
For the ViT methods, all three types of auxiliary tasks that we consider provide a benefit over plain ViT training.
Furthermore, ViT reconstruction-based tasks are found to significantly outperform ViT contrastive-learning.
\end{abstract}


\section{Introduction}



Image-based reinforcement learning is an important and growing area, 
as  cameras provide a ubiquitous and inexpensive way to acquire observations in complex and unstructured environments.
However, in RL, it remains common to give privileged access to compact state descriptors that may not be available in real-world settings.
This is because extending such methods to work with images is non-trivial, usually requiring 
significant additional computation and algorithmic improvements to cope with the high-dimensional nature of images as inputs.
In recent years, end-to-end deep reinforcement learning from visual inputs has yielded impressive results in domains such as robotics control tasks~\cite{levine2016end, 2019-CORL-cassie}, Atari games~\cite{mnih2015human} and autonomous driving~\cite{kendall2019learning}.
However, such methods remain data-intensive and are brittle with respect to confounding visual factors, including dynamic backgrounds, other agents, and changing camera perspectives.

In principle, end-to-end RL can learn representations directly while learning the policy. 
However, prior work has observed that RL is bounded by a "representation learning bottleneck" in the sense that a considerable portion of the learning period must be spent acquiring good representations of the observation space.
To mitigate this, the learning of good state representations can be aided by auxiliary losses that guide the learning of a suitable  representation~\cite{hawke2020urban, yarats2019improving}.
They can also be learned fully in advance, in an unsupervised fashion, in support of the subsequent control tasks to be learned~\cite{ha2018recurrent, higgins2017darla, nair2018visual}.
Recently, data augmentation techniques, already popular in computer vision, have shown to significantly improve performances in RL from pixels~\cite{yarats2020image, yarats2021image, srinivas2020curl, laskin2020reinforcement}.

More recently, computer vision is seeing a potential shift in network architectures from convolutional neural networks (CNNs) to vision transformers (ViT), as the latter are being repeatedly shown to learn good representations for the downstream tasks. Recent advances in Transformers~\cite{vaswani2017transformer} and ViT~\cite{dosovitskiy2020vit} raise the obvious question as to whether or not this type of architecture will also benefit image-based deep reinforcement learning. Since ViT usually requires significantly more data to train on, a common methodology is to first train using a self-supervised objective and then fine-tune the representation as needed for  downstream tasks. Motivated by this approach, we investigate whether ViT assisted with the existing self-supervised training objectives can assist in learning image-based RL policies. 

Our primary contributions are as follows:
\begin{itemize}[leftmargin=\parindent]
    \item We adapt and implement three existing self-supervised learning methods for computer vision tasks as auxiliary tasks 
    for ViT-based RL policies;
    \item We evaluate and compare the above ViT-based appraoaches with RAD, a leading CNN-based RL method.
       Our results point to better overall performance for RAD. For the ViT methods, we find that masking-based methods
       outperform contrastive learning.
\end{itemize}
\section{Related Work}


Reinforcement learning from pixels has been approached using RL algorithms including DQN~\cite{mnih2015human} and DDPG~\cite{lillicrap2015continuous}.
These algorithms were applied to Atari environments and not more complex dynamical control environments like the locomotion ones in Mujoco~\cite{todorov2012mujoco}, PyBullet~\cite{coumans2021} or Deepmind Control Suite~(DMControl Suite)~\cite{tassa2020dmcontrol}.
These algorithms are not explicitly designed for image-based input, and they learn faster with lower dimensional
state representations.

In recent years, a focus on RL from pixel-based input has narrowed the gap between 
learning from compact state observations and high-dimensional observations, generally using CNNs.
Learning a representation with an auxiliary decoder is known to improve efficiency in learning good representations~\cite{kendall2019learning, yarats2019improving}.
Data augmentation methods have recently been adopted in RL, in contrast to computer vision where this has a longer history.
RAD~\cite{laskin2020reinforcement} proposes to augment images with random cropping, and other augmentations to improve data-efficiency.
DrQ~\cite{yarats2020image} and DrQ-v2~\cite{yarats2021image} propose to augment both the input image and the Q-function. This acts as a regularization technique and allows efficient learning of action-value approximations for images. 
Later, stochastic data augmentation is proposed to further stabilize the training of agent~\cite{hansen2021stabilizing}. 
SUNRISE proposes the use of an ensemble of Bellman back-ups and a novel exploration strategy to enhance the performance~\cite{lee2021sunrise}. 
Predicting future internal representations has also been demonstrated to be an effective objective for RL from pixels~\cite{schwarzer2020data}.
CURL~\cite{srinivas2020curl} uses a contrastive loss~\cite{chen2020simple, van2018representation} to match representations of the same image processed with different augmentations. The contrastive objective for RL is then extended to incorporate temporal prediction~\cite{lee2020predictive,stooke2021decoupling} and curiosity-driven rewards~\cite{nguyen2021sample}. Additionally, model-based approaches learn the latent dynamics corresponding to image-based observations to improve sampling efficiency~\cite{ha2018recurrent, hafner2019learning, hafner2019dream, hafner2020mastering}.

Multiple methods have been proposed to learn self-supervised representations with ViT, and later finetune them for downstream computer vision tasks. Popular ideas include contrasting the visual representations of different augmented views of the images~\cite{he2019moco, chen2021empirical}, discriminating the latent representations~\cite{caron2021emerging, zhou2021ibot}, and predicting targets with masked inputs~\cite{he2021masked, wei2021masked, bao2021beit, baevski2022data2vec}.

In this work, we aim to analyze the usage of ViTs for reinforcement learning from pixels. We seek to understand the extent to which additional self-supervised losses can guide the learning of efficient representations in an RL context, and to compare this to CNN-based methods such as RAD.

\section{Experimental Design}
\label{sec:experimental-design}

\begin{figure*}
  \centering
  \includegraphics[width=0.75\linewidth]{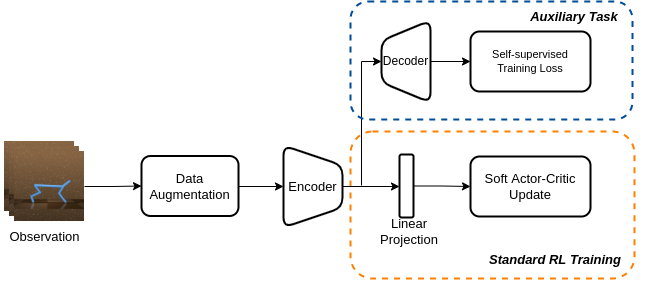}
  \caption{High-level overview of our training pipeline for both ViT and CNN encoders. We keep the standard RL updates unchanged when we switch to ViT auxiliary tasks. We use random cropping as augmentation for RL training. For auxiliary ViT tasks proposed in~\cite{baevski2022data2vec,he2021masked}, we apply random masking as the augmentation stategy, and we adopt another randomly cropped view as augmented sample for contrastive learning~\cite{he2021masked,srinivas2020curl}. }
  \label{fig:overview}
\end{figure*}

Convolutional architectures paired with data augmentation techniques have shown great progress in RL from pixels.
However, whether ViT can bring any benefit to the problem remains unanswered yet. 
Therefore, we study the impact of a plain ViT encoder and various self-supervised losses proposed with ViT on the RL from pixels problem.
We embed three recently proposed self-supervised objectives with ViT into the standard RL pipeline as auxiliary tasks. More specifically, we employ the idea of \emph{Data2Vec}~\cite{baevski2022data2vec},
\emph{MAE}~\cite{he2021masked},
and momentum contrastive learning~\cite{chen2021empirical,srinivas2020curl}. 

A high-level summary of our empirical study is illustrated in~\autoref{fig:overview}.
We keep the standard RL training fixed at all times, and test the performance of adding different ViT auxiliary tasks,
as well as comparing to a CNN-encoder, as implemented for RAD.
In this pipeline, the stacked images are first augmented via random cropping, and then embedded to a shared representation using an encoder.
This shared representation is then used both for RL training, after being flattened and passed through fully-connected layers,
and the auxiliary training tasks, if any.
For ViT, we adopt the ViT architecture as the encoder.
For a fair comparison, we choose a small ViT encoder with a similar order of magnitude of learnable parameters as its convolutional counterpart.

We now introduce each self-supervised objective briefly. Please refer to the original paper for more details~\cite{baevski2022data2vec, he2021masked, chen2021empirical, srinivas2020curl}.

\begin{figure*}
  \centering
  \begin{tabular}{c|c|c}
        \begin{subfigure}{0.29\textwidth}
            \includegraphics[width=\linewidth]{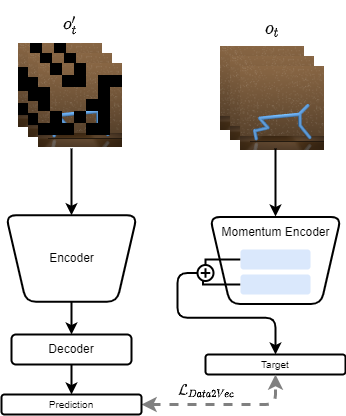}
            \caption{Data2Vec}
            \label{fig:data2vec}
        \end{subfigure}&
        \begin{subfigure}{0.28\textwidth}
            \includegraphics[width=\linewidth]{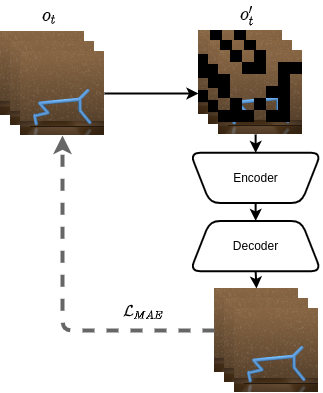}
            \caption{MAE}
            \label{fig:mae}
        \end{subfigure}&
        \begin{subfigure}{0.30\textwidth}
            \includegraphics[width=\linewidth]{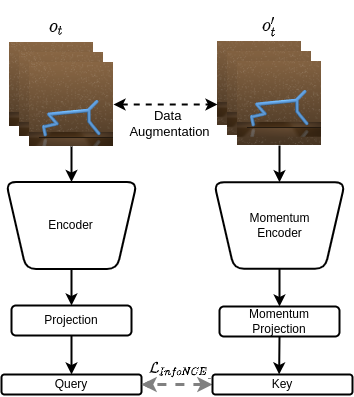}
            \caption{Momentum contrastive learning}
            \label{fig:contrastive}
        \end{subfigure}
  \end{tabular}
 \caption{A detailed representation of the different auxiliary tasks used to guide the learning of the encoder. 
 (a) Data2vec receives two variations of the same image and tries to match the reconstruction. 
 (b) MAE uses an encoder-decoder architecture receiving as input an image with missing patches and tries to reconstruct the original image. 
 (c) The contrastive loss~(InfoNCE) draws together similar image pairs (different augmentations of the same image) while pushing apart dissimilar image pairs.}
 \label{fig:comparison}
\end{figure*}

\subsection{Data2Vec}
\emph{Data2Vec}~\cite{baevski2022data2vec} is a unified self-supervised model proposed across multiple modalities with different specialized structures, including images, speeches and natural language processing. In this work, we focus on its variant operating on images. Data2Vec adopts the idea of predicting the internal representation of a masked input to match the representation of the original input. A visual depiction is shown in \autoref{fig:data2vec}.

The self-supervision task receives the original stack of images~$o_{t}$ and its masked version~$o'_{t}$ as input.
The observation~$o_{t}$ is encoded with a momentum encoder~$E'_{\theta'}$. The output of the momentum encoder summed with the activation value of the $K$-last layers of the encoder forms the target vector~$t$.
Instead, the masked input~$o'_{t}$ is encoded with an encoder~$E_{\theta}$ and then passed through a two-layer fully-connected decoder to form a prediction vector~$p$.
The goal is to minimize the difference between the prediction~$p$ and the target~$t$.

To avoid learning trivial solutions (i.e., degenerating the representation into a uniform vector), the authors apply parameter-free layer normalization on the output value~$a^{i}$ of the last $i^{th}$ layer. The target~$t$ can be mathematically expressed as: $t = \sum_{i=0}^{K} \text{LayerNorm}(a^{i})$

Given the target~$t$ and prediction~$p$, the parameters of the encoder and the decoder are updated through the Smooth L1 loss as follows:
\begin{equation}
    \mathcal{L}_{Data2Vec}(t, p) = 
    \begin{cases}
        \frac{1}{2}(t-p)^{2} \// \beta & \text{if} \|t-p\| \leq \beta,\\
        \|t-p\|-\frac{1}{2}\beta & \text{otherwise.}
    \end{cases}
\end{equation}
The momentum encoder~$E'_{\theta'}$ is updated through Polyak averaging of the encoder~$E_\theta$'s parameters: 
\begin{equation}
    \theta' = (1-\tau) \theta + \tau \theta'.
    \label{eqn:ema}
\end{equation}

\subsection{Masked Autoencoders}
\emph{MAE}~\cite{he2021masked} adopts a simple-yet-effective idea: given an image with masked patches, the self-supervised objective is to reconstruct the original unmasked image. Both the encoder and decoder use a ViT architecture. A visual representation of \emph{MAE} is depicted in \autoref{fig:mae}.

The encoder only operates on the unmasked patches and encodes the unmasked patches to a latent representation~$z_{t}$.
The decoder then receives the concatenation of both the unmasked embeddings and the masked patches as input, and transforms the input into images~$p$.
All the masked patches are uniformly represented by a learnable vector~$z'$.
The positional information of each patch is indicated by the position embedding of the decoder.
We implement a lightweight decoder to save computational resources.
The reconstruction target $t$ is the pixel value of the masked patches only, and the pixels are normalized per patch to improve the performance as the authors of \cite{he2021masked} suggest.
The reconstruction loss~$\mathcal{L}_{MAE}$ is computed as the mean squared error between the target~$t$ and reconstructed images~$p$ only on the masked portion.

\subsection{Momentum Contrastive Learning}

Contrastive learning~\cite{chen2020simple, he2019moco, henaff2020data, wu2018unsupervised} is designed to learn representations that obey similarity constraints in a self-supervised manner. The process of contrastive learning can be understood as a dictionary look-up task. Given an encoded query~$q$ and a dictionary of encoded keys~$K = \{k_{0},k_{1},k_{2},\dots,k_{n}\}$, among which one key~$k^{+}$, defined as the positive key, matches the query~$q$, the objective of contrastive learning is to ensure the distance between the query~$q$ and the positive key~$k^{+}$ is closer than the distance of the query~$q$ and any other key in the dictionary~$K\setminus{k^{+}}$. The keys are commonly encoded via a momentum encoder~\cite{he2019moco,chen2020improved,he2021masked} to enhance training stability. We adopt the loss design of CURL\cite{srinivas2020curl} where the distance is computed as bilinear products~($q^{T}Wk$) with a learnable matrix $W$. InfoNCE is applied as the loss to ensure the similarity constraints:
\begin{equation}
    \mathcal{L}_{\text{InfoNCE}} = \text{log}\frac{\text{exp}(q^{T}Wk_{+})}{\text{exp}(q^{T}Wk_{+}) + \sum_{i=0}^{K-1}\text{exp}(q^{T}Wk_{i})},
\end{equation}

As shown in \autoref{fig:contrastive}, the encoder and momentum encoder compress the two sets of image stacks~$\{o_{t}, o'_{t}\}$ into latent vectors as queries and keys. We treat the image stack yielded from the same image but augmented differently as the positive key~$k^{+}$, and other image stacks in the batch as the negative keys~$K\setminus{k^{+}}$. The momentum encoder is a moving average of the encoder updated according to \autoref{eqn:ema}.

\section{Experiments and Results}

\begin{table*}
  \centering
  \begin{tabular}{lcccc|c}
    \midrule
    \textbf{100K STEP SCORES} & ViT & ViT w/ Data2Vec & ViT w/ MAE & ViT w/ Contrastive & RAD\\
    \midrule
    FINGER, SPIN  & $293.06\pm260.74$ & $646.52\pm247.59$ & \textbf{\textit{899.08$\pm$106.28}} & $850.36\pm103.52$ & $823.16\pm169.13$ \\
    CARTPOLE, SWING  & $852.73\pm15.64$ & $721.91\pm283.46$ & \textit{864.45$\pm$4.90} & $844.48\pm27.11$ & \textbf{871.50$\pm$9.87} \\
    REACHER, EASY  & $125.72\pm60.36$ & $185.98\pm73.35$ & $383.52\pm234.86$ & \textit{440.52$\pm$88.83} & \textbf{897.56$\pm$73.12} \\
    CHEETAH, RUN  & $263.54\pm22.79$ & $287.46\pm36.44$ & \textit{366.78$\pm$49.08} & $270.40\pm147.41$ & \textbf{584.30$\pm$15.83} \\
    WALKER, WALK  & $562.92\pm105.89$ & \textit{587.84$\pm$101.82} & $263.57\pm228.29$ & $377.67\pm258.46$ & \textbf{875.04$\pm$138.11} \\
    CUP, CATCH  & $172.06\pm182.79$ & $410.50\pm192.20$ & \textbf{\textit{946.34$\pm$21.87}} & $804.28\pm235.83$ & $661.63\pm140.37$ \\
    \midrule
    \textbf{500K STEP SCORES} &  & &  &  & \\
    \midrule
    FINGER, SPIN  & $758.82\pm381.72$ & \textbf{\textit{975.54$\pm$8.80}} & $951.62\pm53.20$ & $917.80\pm127.19$ & $843.86\pm166.31$ \\
    CARTPOLE, SWING  & $861.76\pm14.02$ & $866.74\pm12.31$ & \textit{868.34$\pm$10.96} & $853.49\pm32.11$ & \textbf{872.03$\pm$7.23} \\
    REACHER, EASY  & $230.08\pm77.42$ & $234.28\pm91.76$ & $362.04\pm83.65$ & \textit{539.70$\pm$119.56} & \textbf{910.26$\pm$50.58} \\
    CHEETAH, RUN  & $480.15\pm50.48$ & $539.86\pm104.35$ & \textit{551.58$\pm$50.48} & $442.82\pm96.81$ & \textbf{837.36$\pm$26.83} \\
    WALKER, WALK  & $841.77\pm48.30$ & \textit{895.56$\pm$37.94} & $844.29\pm43.90$ & $892.41\pm78.93$ & \textbf{970.20$\pm$7.08} \\
    CUP, CATCH  & $486.34\pm394.58$ & $963.04\pm4.82$ & \textbf{\textit{973.64$\pm$5.78}} & $888.18\pm132.07$ & $925.16\pm20.84$ \\
    \bottomrule
  \end{tabular}
  \caption{Average test reward for the environments in DMControl Suite at 100K and 500K training steps. Method achieving the best performance is highlighted in bold text while the best performance among the ViT variants is labelled in italic text.
  }
  \label{tab:reward}
\end{table*}
We evaluate different choices of self-supervised learning losses, including \emph{Data2Vec}~\cite{baevski2022data2vec}, \emph{MAE}~\cite{he2021masked}, and contrastive learning approaches~\cite{srinivas2020curl,he2019moco} with ViT on continuous control tasks in DMControl Suite~\cite{tassa2020dmcontrol}. We choose a vanilla ViT pipeline without any auxiliary task as the baseline for comparison, and also report the results for RAD~\cite{laskin2020reinforcement}, the state-of-the-art~(SOTA) convolutional architecture for RL from pixels for a complete view. We keep the number of parameters as similar as possible across all the experiments. To make our comparisons as fair as possible, we keep the following details fixed in all ViT-based experiments: (i) Soft Actor-Critic (SAC) as RL algorithm, (ii) structure and dimensionality of ViT encoders, (iii) learning rate for auxiliary tasks and RL updates.
We use hyperparameter values from the respective original papers and keep hyperparameters unchanged across the experiments, except for the masking ratio used in~\cite{he2021masked,he2019moco}. Random cropping is selected as the image augmentation strategy for all the experiments in the standard RL training. Further implementation and hyperparameter details are described in Appendix~\ref{app:implementation-details}.

For evaluation, we use the average total reward at 100K and 500K gradient update scores, as listed in \autoref{tab:reward}.
Each entry is evaluated by averaging across 5 runs with different random seeds.
Corresponding learning curves are illustrated in \autoref{fig:learning_curve}.
Compared with a plain ViT model, adding any auxiliary task improves the learning performance. 
Among the three auxiliary training tasks, the reconstruction-based tasks (\emph{Data2Vec} and \emph{MAE}) are clearly better than the contrastive learning one. \emph{MAE} outperforms \emph{Data2Vec} in most environments except the \texttt{WALKER,WALK} environment.
When convolutional models~(\emph{RAD}) are also considered for comparison, \emph{RAD} still achieves the SOTA performance in 4 out of the 6 experiments, while \emph{Data2Vec} and \emph{MAE} with ViT architecture reach the best accumulated reward in the \texttt{FINGER, SPIN} and \texttt{CUP,CATCH} environments.

\section{Discussion and Future Work}


Adding reconstruction-based auxiliary tasks~\cite{baevski2022data2vec,he2021masked} with ViT encoders can significantly enhance performance while encoders built with CNNs still obtain the best performance in the majority of our experiments. 
\emph{Data2Vec} and \emph{MAE} both predict the features of a masked version of the input. \emph{Data2Vec} reconstructs the deep features encoded by the momentum encoder while \emph{MAE} chooses the original pixel values as the reconstruction target. 
The contrastive learning framework improves the performance of a plain ViT to a limited extent, which is analogous to the study on CNN~\cite{laskin2020reinforcement,srinivas2020curl}. 
We suspect that a lack of diversity in the training images could cause the inferior performance of contrastive learning. Unlike the diverse \emph{ImageNet} dataset often used as the baseline for computer vision research, images collected by the RL agent are more similar to each other. As a result, positive and negative keys are sometimes too similar to be distinguished for the contrastive objective. 

Since transformer-based architectures are notorious for being data-hungry, we are also interested in further improving the sampling efficiency via pre-training. \emph{First pre-train, then finetune} is a well-established paradigm for learning computer vision models to mitigate the issue of insufficient data. Due to  a lack of datasets for control tasks, and more specifically RL, what serves as the best dataset for pre-training remains an open question. We plan to investigate the following options: (i) a large, general and diverse dataset such as ImageNet, (ii) a dataset generated from an expert policy, (iii) a dataset generated from a random policy, and (iv) a dataset containing a mixture of expert and non-expert demonstrations.

\section{Conclusion}

This paper empirically explored the potential of several ViT methods for RL and compared the results to CNNs trained with RAD, a leading method for image-based RL for CNNs. 
Specifically, we evaluate vanilla ViTs, as well as the addition of various self-supervised losses (reconstructive and contrastive) to ViT as auxiliary tasks to the RL training. 
Our results show that ViTs trained with auxiliary tasks (self-supervised losses) are helpful for the RL-from-pixels problem, 
but they currently still fall short of what can be achieved with CNNs by RAD.

\appendix
\begin{figure*}[h!]
    \centering
    \begin{subfigure}{0.3\textwidth}
        \includegraphics[width=\linewidth]{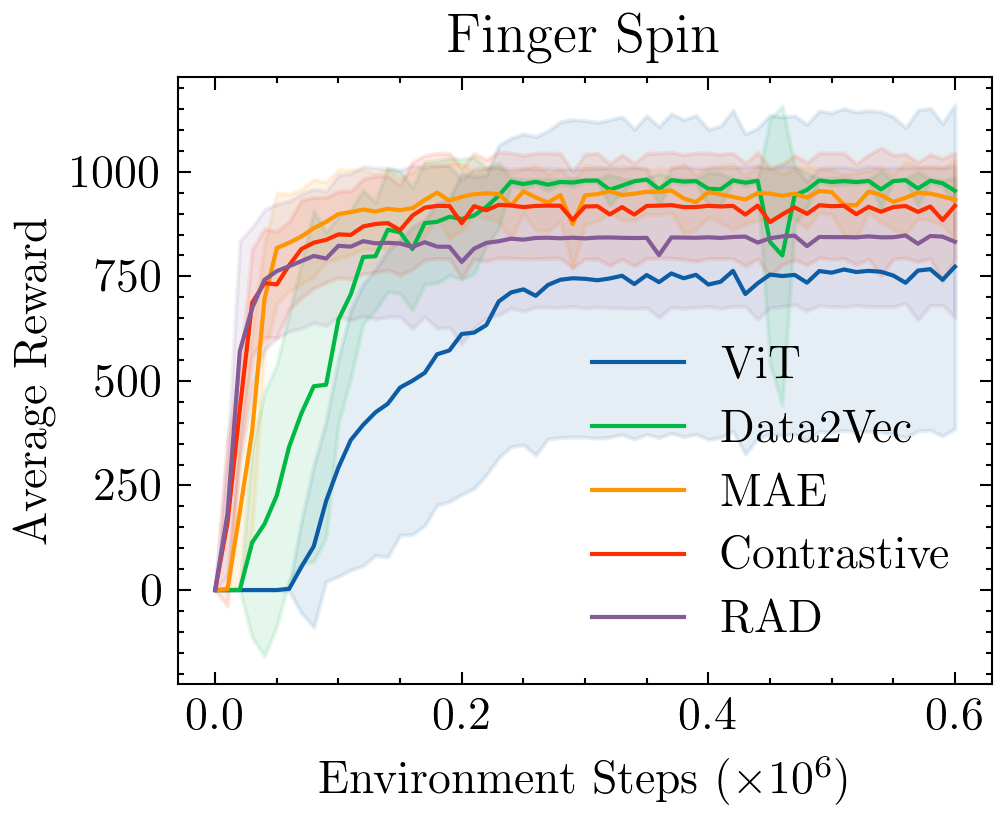}
    \end{subfigure}
    \begin{subfigure}{0.3\textwidth}
        \includegraphics[width=\linewidth]{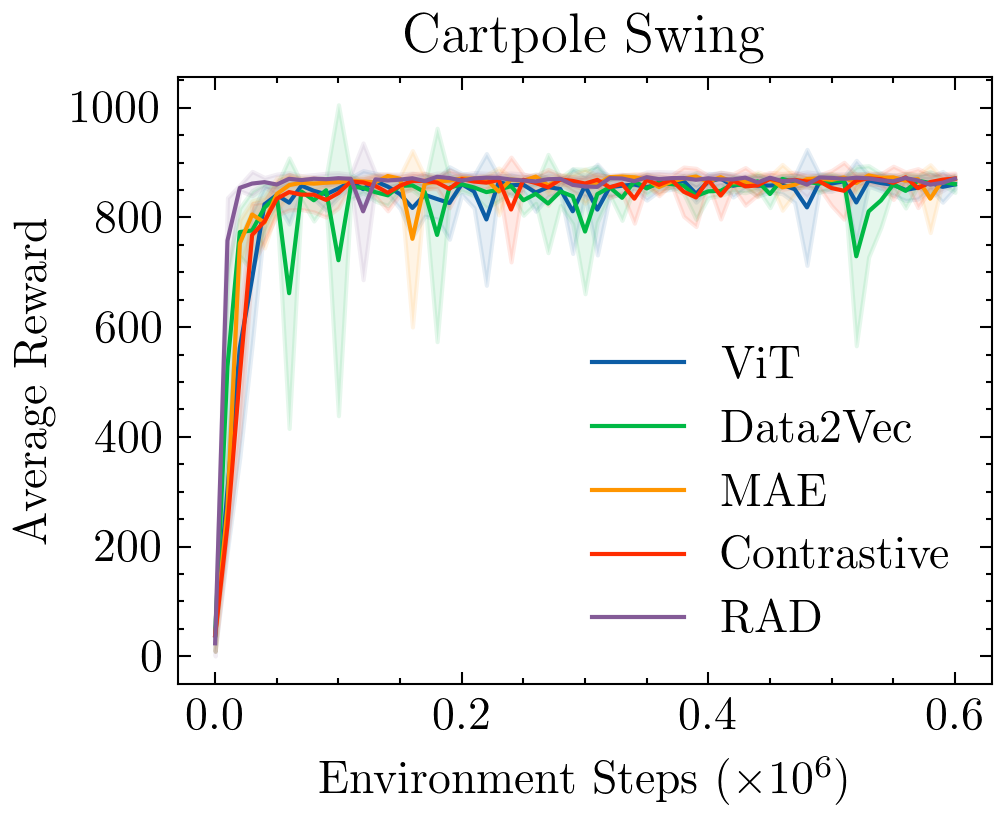}
    \end{subfigure}
    \begin{subfigure}{0.3\textwidth}
        \includegraphics[width=\linewidth]{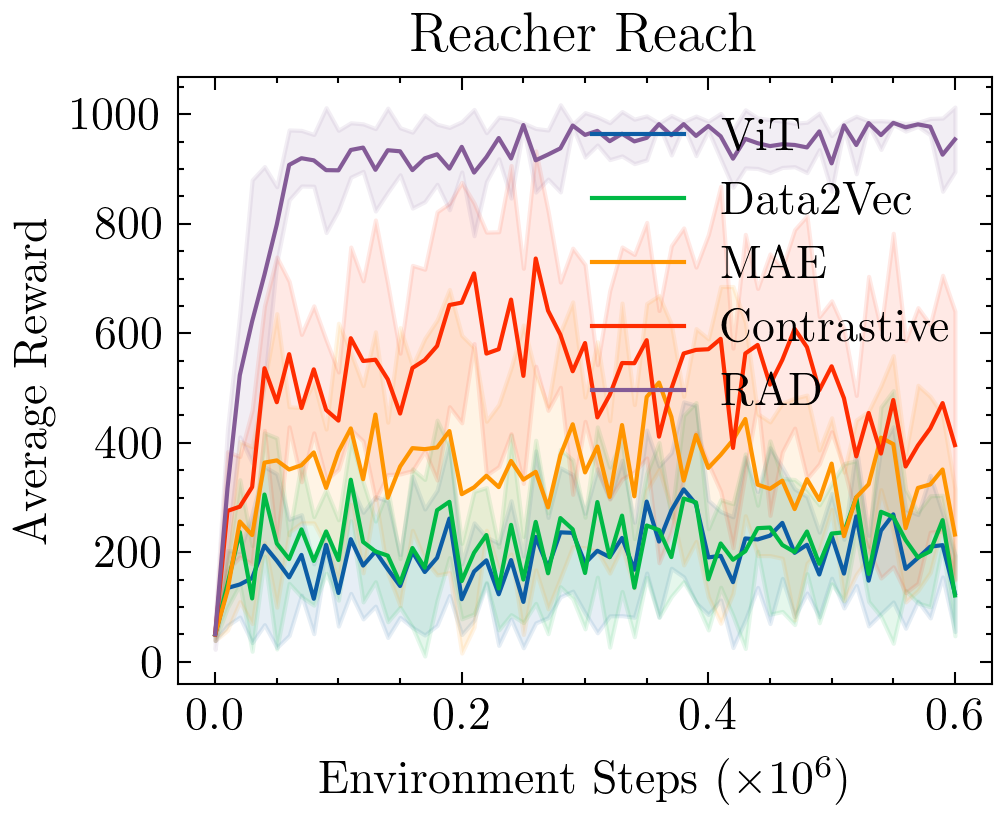}
    \end{subfigure}
    \begin{subfigure}{0.3\textwidth}
        \includegraphics[width=\linewidth]{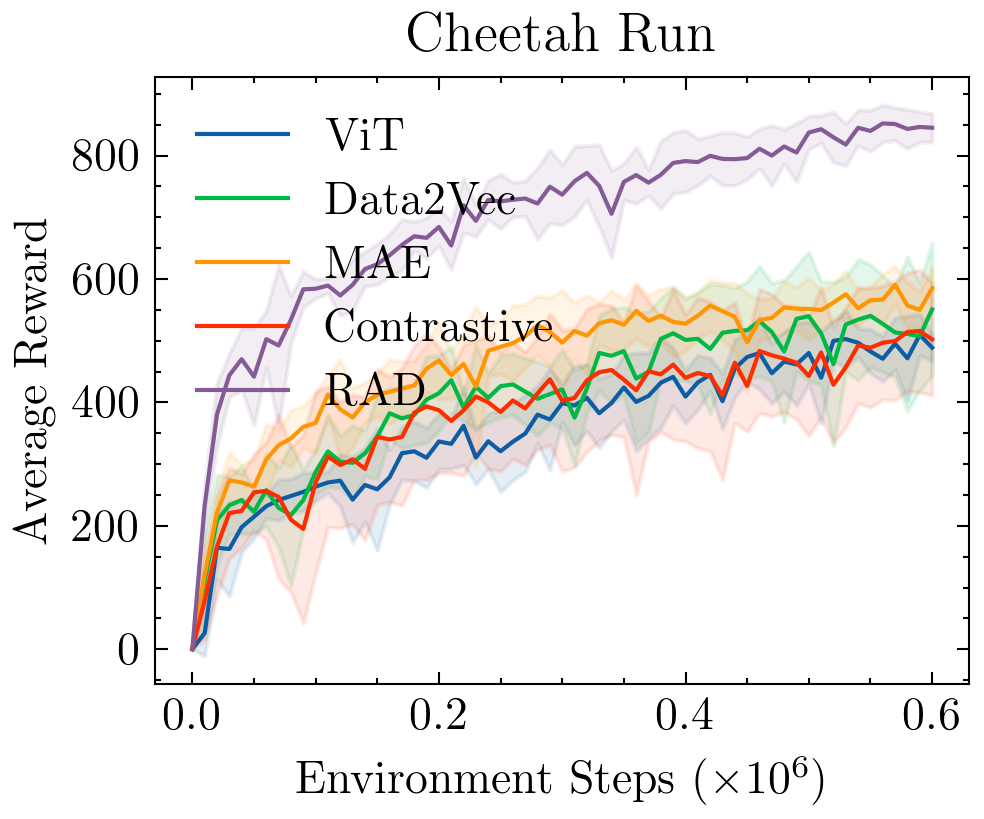}
    \end{subfigure}
    \begin{subfigure}{0.3\textwidth}
        \includegraphics[width=\linewidth]{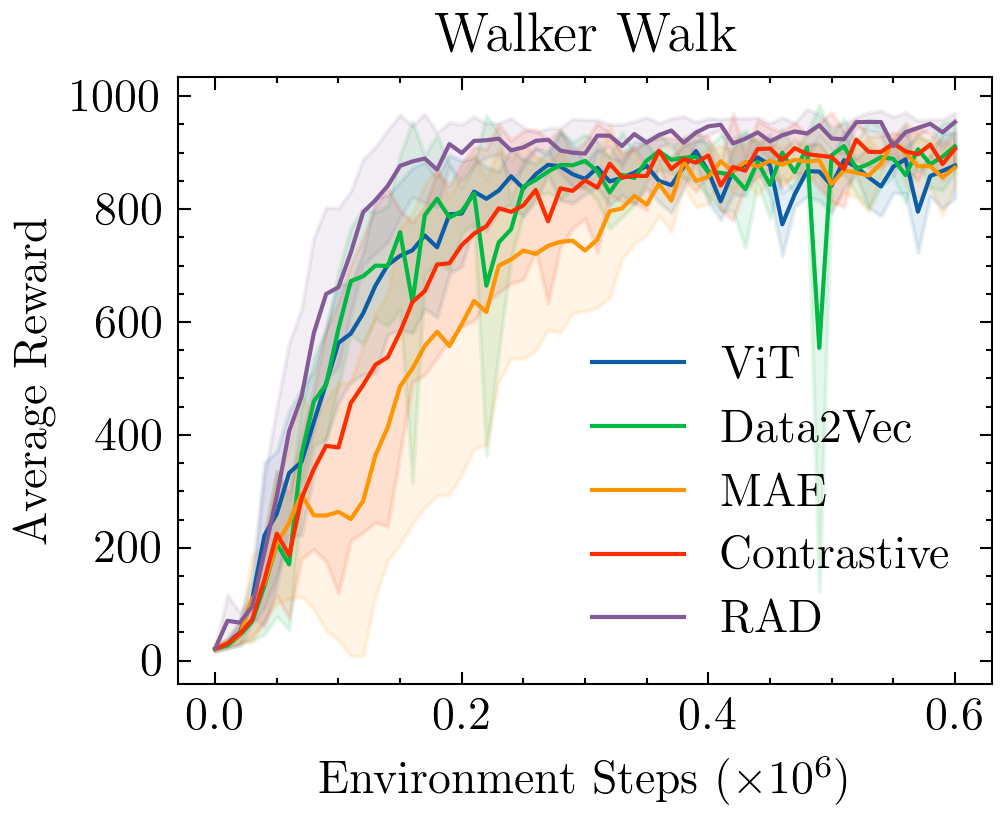}
    \end{subfigure}
    \begin{subfigure}{0.3\textwidth}
        \includegraphics[width=\linewidth]{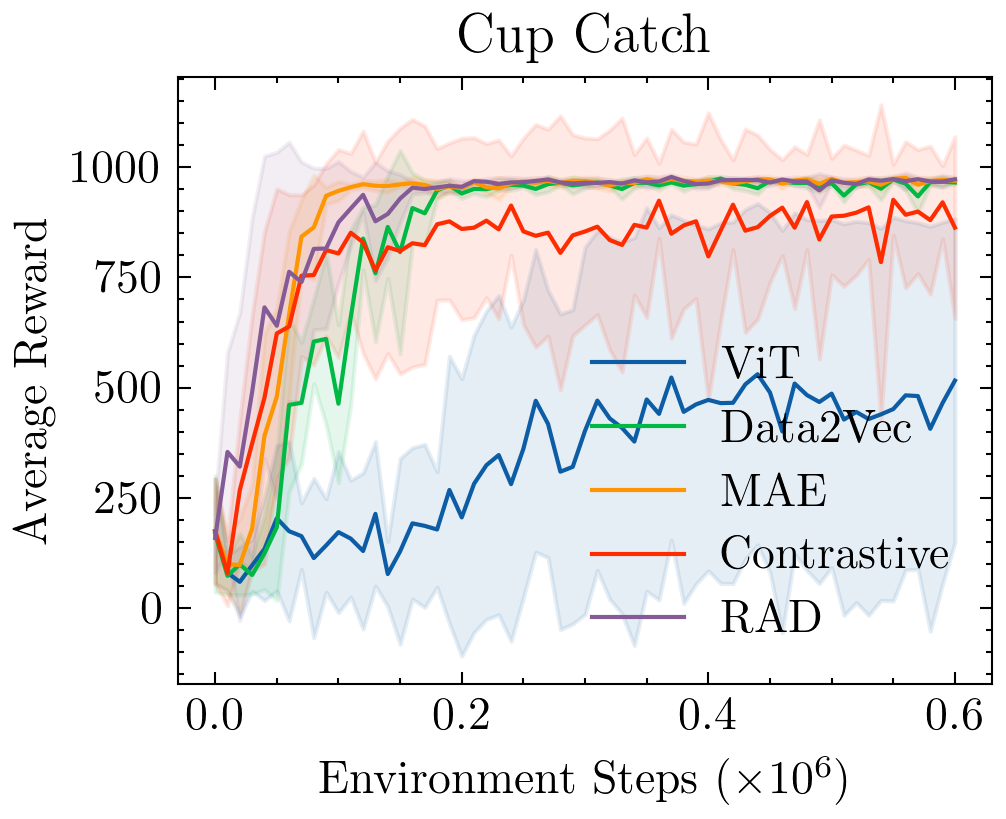}
    \end{subfigure}
 \caption{Learning curves of Data2Vec, MAE and contrastive learning on DMControl Suite.}
 \label{fig:learning_curve}
\end{figure*}
\subsection{Background}
\label{app:background}
\subsubsection{Reinforcement Learning}

We formulate the RL problem as a Markov Decision Process~(MDP). An MDP consists of the following elements: states $s_{t} \in \mathcal{S}$, actions $a_{t} \in \mathcal{A}$, a dynamics function $p(s_{t+1}|s_{t}, a_{t})$ denoting the transition probability of a state-action combination~$(s_{t}, a_{t})$ and a reward function~$R(s_{t}, a_{t})$. Model-free reinforcement learning algorithm often targets at maximizing the expected return as $J = \sum_{t=0}^{T} \gamma^{t} R(s_{t}, a_{t})$ given a discount factor~$\gamma \in [0,1]$. The state variable for RL from pixels problems are commonly represented as a stack of consecutive image frames to infer the status.

Soft Actor-Critic~(SAC)~\cite{haarnoja2018soft} is applied widely to image-based RL problems because of its excellent sampling efficiency and exploration strategy. SAC learns a policy network~$\pi_{\theta}(s_{t})$ and a critic network~$Q_{\phi}^{\pi}(s_{t},a_{t})$ by optimizing the expected return and an entropy regularization term concurrently. The parameters in the critic~$\phi$ are updated by minimizing the squared Bellman error given the transition tuples~$\tau_{t}=(s_{t}, a_{t}, s_{t+1}, r_{t})$ stored in the replay buffer~$D$:
\begin{equation}
\begin{aligned}
    \mathcal{L}(\phi) &= \mathbb{E}_{\tau \sim D}[Q_{\phi}(s_{t}, a_{t}) - (r_{t} + y(r_{t}, s_{t+1}))]^{2},\\
    y(r_{t}, s_{t+1}) &= \gamma(\min_{i=1,2}Q'_{\phi_{i}}(s_{t+1}, a')-\alpha \log_{\pi_{\theta}}(a'|s_{t+1}))
\end{aligned}
\end{equation}
The parameter $\alpha$ is the temperature value to balance the two terms, which is treated as learned parameter as in~\cite{haarnoja2018softapplications}. $Q'$ is a slowly updated copy of the critic to improve training stability. The actor is learned by maximizing the weighted objective of the expected return and the policy entropy as:
\begin{equation}
    \mathcal{L}(\theta) = \mathbb{E}_{a \sim \pi}[Q_{\phi}(s_{t}, a) - \alpha \log_{\pi_{\theta}}(a|s_{t})].
\end{equation}

\subsubsection{Data Augmentation in Reinforcement Learning}

Data augmentation has been found to be essential for the performance of RL from pixels. Common data augmentation techniques include random cropping, color jittering, flipping, rotating and gray scaling. We refer the reader to a detailed study for the comparison between different data augmentation strategies~\cite{laskin2020reinforcement}.
We adopt the most effective random cropping strategy to the images for all the experiments in this work.
Intuitively, random cropping enhances translational invariance to the perception module.
In our experiments, the rendered images have 100 $\times$ 100 pixels, which are later randomly cropped to 84 $\times$ 84 pixels.

\subsubsection{Vision Transformer}

The transformer architecture~\cite{vaswani2017transformer} raised the bar in most domains of machine learning, and have become the state of the art method in NLP tasks.
The Vision Transformer (ViT)~\cite{dosovitskiy2020vit} instead takes the transformer architecture and adapts it to make it suitable for images, and has shown incredible results at classification challenges.
ViT operates by subdividing an image into fixed-size patches. Each patch is then flattened and linearly transformed into an embedding which is concatenated to a learnable 1D positional embeddings to allow for spatial awareness between the input patches. This embedding is then finally processed through the standard transformer architecture.

More recently, ViTs have been used with self-supervised learning to learn good representations, examples of which are Data2Vec~\cite{baevski2022data2vec}, Masked Autoencoders~\cite{he2021masked}. These works apply transformations to the unlabelled images and use losses that try to draw closer together representation of inputs coming from the same original image. A more complete explanation on these methods is described in~\autoref{sec:experimental-design}.

\subsection{Implementation Details}
\label{app:implementation-details}

We adopt the soft actor-critic implementation offered by \cite{pytorch_sac}, which tunes the temperature automatically with a constrained optimization~\cite{haarnoja2018softapplications}. We list the hyperparameters in Table.~\ref{tab:hyper}. For experiments using masked images as input, we experiment with masking ratio from $\{30\%, 40\%, 50\%, 60\%, 75\%\}$, and select the $40\%$ and $75\%$ for \emph{Data2Vec} and \emph{MAE} respectively. The mask is represented as a learnable vector optimized by the auxiliary task. We also find that selecting a small batch size for the contrastive learning can stabilize the training. Therefore, the batch size is 512 for the SAC update but 128 for the contrastive objective.

In our experiments, the policy is trained jointly with the auxiliary tasks. We adopt the training setup from \emph{SAC+AE}~\cite{yarats2019improving}. \emph{SAC+AE}~\cite{yarats2019improving} blocks the gradient signals from the actor to update the shared encoder while the critic has the privilege to update the encoder, which has been proven to greatly improve the performance. 

\begin{table}
  \centering
  \begin{tabular}{ll}
    \toprule
    \textbf{Hyperparameter} & \textbf{Value} \\
    \midrule
    Observation rendering & (100, 100) \\
    Observation downsampling & (84, 84) (random cropping)\\
    Replay buffer size & 100000 \\
    Initial steps & 1000 \\
    Stacked frames & 3 \\
    \multirow{3}{*}{Action repeat} & 2 FINGER, SPIN; WALKER, WALK \\
                                   & 8 CARTPOLE, SWING \\
                                   & 4 otherwise \\
    SAC hidden units(MLP) & 1024 \\
    Evaluation episodes & 10 \\
    Evaluation frequency & 10000 \\
    Optimizer & Adam \\
    Encoder learning rate & $1e-3$ \\
    Actor learning rate & $1e-3$ \\
    Critic learning rate & $1e-3$ \\
    Temperature learning rate & $1e-4$ \\
    Batch Size & 512 \\
    Encoder EMA $\tau$ & 0.05 \\
    Critic function EMA $\tau$ & 0.01 \\
    Discount factor $\gamma$ & 0.99 \\
    Initial temperature & 0.1 \\
    Latent dimension & 128 \\
    Critic update frequency & 2 \\
    Patch size & (12, 12) \\
    ViT depth & 4 \\
    ViT MLP dimension & 128 \\
    Attention head & 8 \\
    K in \emph{Data2Vec} & 2 \\
    $\beta$ in \emph{Data2Vec} & 2.0 \\
    \bottomrule
  \end{tabular}
  \caption{Hyperparameters used in the experiments.
  }
  \label{tab:hyper}
\end{table}

The decoder in \emph{Data2Vec} consists of two-layer MLPs with ReLU activation function. We choose a light-weight ViT decoder for \emph{MAE}, which comprises 2 layers with 64 hidden units and 4 heads. Similar to the conclusion in \emph{MAE}, the capacity of the decoder has limited influence on the overall performance. In our implementation of momentum contrastive learning, the queries and keys are generated by a separate head other than the one used for RL, which can be viewed as the decoder in~\autoref{fig:overview}.
\subsection{Extra Discussion on Image Augmentation}
\label{app:image-aug}

In our work, we apply random cropping to all the experiments as the augmentation technique. The impact of data augmentation with ViT can be further investigated. A corresponding study is conducted with CNN~\cite{laskin2020reinforcement}. Since ViT operates on image patches, diverse augmentation strategies~\cite{cubukpractical,zhang2017mixup, yun2019cutmix, zhong2020random, hoffer2020augment} have been developed to improve training ViT for computer vision tasks. Finding the most effective augmentation technique for RL with ViT is still of great interest to the RL researchers.

\bibliographystyle{IEEEtran}
\bibliography{IEEEabrv, reference}

\end{document}